\pdfoutput=1

\documentclass[11pt]{article}

\usepackage[]{acl}

\usepackage{times}
\usepackage{latexsym}

\usepackage{graphicx} 
\usepackage{booktabs} 
\usepackage{lipsum} 
\usepackage{stfloats}
\usepackage{tabularx}
\usepackage{soul}
\usepackage{adjustbox}
\usepackage{appendix}
\usepackage{ctable}
\usepackage{amssymb}
\usepackage{mathtools, nccmath}
\usepackage{amsmath}
\usepackage{stmaryrd}  
\usepackage{caption}
\usepackage{subcaption}
\usepackage{multirow}
\usepackage{epstopdf}
\usepackage{paralist}
\usepackage[utf8]{inputenc}
\usepackage{multirow}
\usepackage{xstring}
\usepackage[T1]{fontenc}

\usepackage{xstring}

\usepackage[T1]{fontenc}

\usepackage[utf8]{inputenc}

\usepackage{microtype}

\usepackage{inconsolata}

\usepackage{booktabs, makecell, tabularx}

\usepackage{siunitx}
\usepackage{graphicx}
\usepackage{xcolor}

\title{The GPT-WritingPrompts Dataset: A Comparative Analysis of Character Portrayal in Short Stories}

\author{Xi Yu Huang$^{1,3}$ \and Krishnapriya Vishnubhotla$^{1,3}$ \and Frank Rudzicz$^{2,3}$ \\
        \textsuperscript{1}Department of Computer Science, University of Toronto \\ \textsuperscript{2}Faculty of Computer Science, Dalhousie University \\ \textsuperscript{3}Vector Institute, Toronto, ON, Canada \\ \texttt{xiyu.huang@mail.utoronto.ca  vkpriya@cs.toronto.edu frank@dal.ca}
        }
        
\begin{document}
\maketitle
\begin{abstract}
The improved generative capabilities of large language models have made them a powerful tool for creative writing and storytelling. It is therefore important to quantitatively understand the nature of generated stories, and how they differ from human storytelling. 
We create the GPT-WritingPrompts dataset, which pairs short stories written by Reddit users in response to 97,219 prompts with comparable generations from GPT-3.5.
We quantify the emotional and descriptive features of storytelling for both generative processes, human and machine, along a set of six dimensions. We find that generated stories differ significantly from human stories along all six dimensions,
and that human and machine generations display similar biases when grouped according to the narrative point-of-view and gender of the main protagonist. 
We make our dataset and code publicly available\footnote{\url{https://github.com/KristinHuangg/gpt-writing-prompts}}.
\end{abstract}

\section{Introduction}

Storytelling is fundamentally human, and a rich representation of the beliefs, practices, and languages of communities.
The domain of literature is therefore a  source of cultural information for researchers in the social sciences, from historians to linguists \cite{piper2021narrative, ziems2024can}. 
The characters\footnote{We use the term `character' in this work to refer to fictional people, i.e, named entities that speak and act in stories.
} in a story are often central to this process: protagonists are usually depicted with socially and personally desirable traits like kindness, intelligence, and strength, and villains with undesirable characteristics (greed, cruelty, apathy, weakness). 
Consequently, we have seen extensive research on NLP methods that can model and understand these aspects of stories, and draw inferences about social biases represented in them \cite{fast2016shirtless, Lucy2021GenderAR, sap2022quantifying}.

Large language models (LLMs), pretrained on billions of tokens of text from the internet, are now able to generate several thousand words of coherent output in response to user-specified input contexts \cite{bommasani2021opportunities}.
Naturally, these systems have become popular tools for a number of tasks involving text generation, both controlled and open-ended. e.g., dialogue systems, code generation, summarization, question answering, and creative writing. However, neural models have been shown to replicate and amplify the biases encoded in their training data \cite{bender2021dangers}. It is therefore interesting -- and important --  to understand the nature of stories that are told by these models, and how they differ from human models of storytelling \cite{chakrabarty2023art, xie-etal-2023-next, chakrabarty2024creativity, breithaupt2024humans}.


Past work in this area has examined stories that are generated in response to prompts sampled from contemporary fiction novels \cite{Lucy2021GenderAR}, or highly-structured short-story datasets \cite{Mostafazadeh2016ACA, Huang2021UncoveringIG}, for differences in the power, agency, appearance, and intellect attributed to the characters. We contribute to this area of research by:\\[-18pt]
\begin{enumerate}
\setlength{\itemsep}{0pt}
    \item Extending the Reddit WritingPrompts dataset \cite{Fan2018HierarchicalNS} to include generations from the OpenAI GPT-3.5 model (\texttt{gpt-3.5-turbo}).  
    The original dataset consists of 97,222 unique writing prompts, and multiple stories written by Reddit users in response to each prompt. We generate a corresponding set of artificial stories by prompting \texttt{gpt-3.5-turbo} with a  role matching that of the human writers, and for a comparable length ($\sim$500 words). The resulting dataset,  \textbf{GPT-WritingPrompts}, is a useful source of free-form natural language short stories with matching human and machine responses to the same prompt. \\[-18pt]
    \item Analyzing the narrative characteristics of stories written by both sets of writers along 
    six dimensions.
    The first three are the affective dimensions of valence (positive--negative), arousal (active--passive), and dominance (powerful--weak), which capture the three fundamental dimensions of connotative meaning as formulated by \citet{Osgood1958TheMO}. We also look at three other axes of character portrayal that are often noted in prior work for stereotypical connotations along gender lines: intellect, appearance, and power.\\[-18pt]
    \item Quantifying biases in character portrayal along these dimensions, in both human-written and GPT-generated stories, when grouped by the narrative voice and the gender of the main protagonist of the story. Stories are categorized as being in first-person, second-person, or third-person, based on the dominant pronouns used in them. Third-person stories are further categorized as featuring a male or female protagonist (or Other, if pronouns do not fall exclusively into one of these categories).\\[-18pt]
\end{enumerate}

We show that GPT-generated stories differ significantly from human-written ones along {\em all} six dimensions, when compared at both the aggregate-level (Sections \ref{res-pov-agg}) and at the instance-level (Section \ref{res-instance-level}). On average, GPT-generated stories use more words with a positive valence and higher dominance compared to human stories, but lower arousal (less activity), and use fewer words that relate to appearance and intellect. Both sets of stories, however, display similar biases towards their male and female protagonists: female protagonists are more positive, less aroused, less in control, are associated with a higher proportion of appearance-related words, and a lower proportion of intellect-related words compared to male protagonists. This supports prior findings on how neural models of language generation differ from those of humans \cite{Lucy2021GenderAR, Huang2021UncoveringIG, giulianelli-etal-2023-comes, breithaupt2024humans}.
\section{Background}

\subsection{Generative Models of Natural Language}
Text generation, both conditional and open-ended, has been a core area of NLP research for many decades. The development of neural transformer models, pretrained with the language modeling objective on large amounts of textual data, has led to exponential improvements in the fluency, coherence, and versatility of machine-generated text \cite{Li2022PretrainedLM}. Popular models include the GPT family of models from OpenAI\footnote{\url{https://platform.openai.com/docs/models}}, Llama models from Meta AI \cite{Touvron2023LLaMAOA}, and the Claude family of models from Anthropic AI \cite{TheC3}, among others. These models are able to generate long-form text (up to thousands of tokens in length) in response to a user-provided input prompt, which has led to their adaptation as the base models for several NLP tasks, including translation, question answering, summarization, and controlled text generation \cite{brown2020language}. 

Parallel to these advances are studies of the probabilistic algorithms behind generative processes of these models, and how they differ from human language generation. Given a context, this involves designing metrics that compare how similar machine generation is to a set of reference human generations (reference-based evaluation). 
Another approach, closer to ours and more appropriate for open-ended text generation, is to study the statistical distributions of various text features (type-to-token ratio, word and sentence lengths) in the two generation processes (humans and machine). These measures can be computed and compared both at the corpus-level, and at the instance-level \cite{meister2021language, pillutla2021mauve, giulianelli-etal-2023-comes}. 

Another equally important area of research is to quantify the \textit{default} generative behaviours of these models, understand the biases exhibited by them towards or against particular cultures, ideologies, or products, and whether they reinforce harmful stereotypes about certain groups, particularly those that have historically been marginalized. 

\subsection{Biases in Stories}
Research into biases in stories, both human and artificial, has focused largely on differences in the portrayal of characters belonging to different demographic groups. 
Studies on the portrayal of gendered groups in particular, though mostly limited to the male/female binary, have demonstrated that female characters tend to be portrayed as being more emotional \cite{fast2016shirtless, zhou2022moraleventcentricinspection}, having less agency and power \cite{sap-etal-2017-connotation}, and are more prone to being described by physical attributes \cite{Lucy2021GenderAR, Huang2021UncoveringIG, begus2023experimentalnarrativescomparisonhuman, Zhao2024ACS}. 

A common strategy for measuring such biases is to first extract descriptive lexical attributes associated with gendered characters from the story text, and then quantify the bias of these terms towards a certain lexical dimension, such as power or appearance, using an associated lexicon.
\citet{fast2016shirtless} create lexicons of words associated with several categories of behavioural stereotypes, such as violent, intelligent, domestic, emotional, etc. \citet{sap-etal-2017-connotation} create lexicons of verbs that quantify the distribution of power and agency between the subject (agent) and object (theme) of verbs. Emotion lexicons have been created for many
emotion dimensions, and in multiple languages, such as LIWC \cite{pennebaker2001linguistic}, WordNet-Affect \cite{bobicev2010emotions}, SentiWordNet \cite{baccianella2010sentiwordnet},
VADER \cite{hutto2014vader}, and the NRC suite of emotion and affect lexicons
\cite{mohammad-2018-obtaining}.

Here, we analyze appearance, intelligence, and power as aspects of character portrayal. We also include the three affective dimensions of valence, arousal, and dominance. We compare and contrast the results of characterizing a protagonist by explicit linguistic markers (verbs and adjectives), as well as commonsense inferences with COMET.

\section{The GPT-WritingPrompts Dataset}

\citet{Fan2018HierarchicalNS} collected  300K stories written in response to 97,223  prompts from the Reddit \texttt{r/WritingPrompts} forum, referred to as the \textbf{WritingPrompts} dataset. Users of the forum post short prompts designed to invoke a  scenario or set the premise for a  narrative to unfold, and other forum participants respond with short stories relating to, or following, the premise set-up in the prompt. 

Appendix Table \ref{tab:prompt-story-example} shows a sample of three prompts and randomly-sampled submitted story for each. As we can see, the prompts and stories evoke a wide-ranging set of topics (apocalyptic war scenarios, everyday life with a twist, historic settings). The style of the stories also demonstrates a lot of variety: some are in third-person, with one or more explicitly-referenced protagonists, and others are written from a first-person point-of-view. 

We collect all prompts and associated stories in the training set of WritingPrompts. 
Each prompt is associated on average with 2.8 stories, and the stories average 675 tokens in length (Table \ref{tab:basic-stats}). 

\subsection{Generating Artificial Stories}
We now obtain artificial, or machine-generated, stories for each prompt by using a pretrained large language model. Specifically, we prompt the \texttt{gpt-turbo-3.5} model from OpenAI to generate multiple continuations for a given story prompt, slightly varying the user role specified along with the prompt context. 

GPT models allow us to specify a \textit{system role} that helps set the behaviour of the generative model. These roles influence the ``persona" assumed by the model as it generates text for subsequent input contexts -- one can specify a certain style of talking, or a level of intelligence, or formatting specifications for the output. Considering that our domain of narratives is quite specific -- short stories written on an internet forum -- we test two variations of the system role:
\\[-20pt]
\begin{itemize}
    \item \textbf{Author role:} ``You are an award winning creative short story writer."\\[-20pt]
    \item \textbf{Redditor role:} ``You're writing a Reddit story and you want other reddit users to like and upvote your story."
\end{itemize}
An initial, qualitative examination of generated stories for a random sample of 100 prompts did not indicate any obvious stylistic differences between the two system roles; we therefore stick with the Redditor role for the rest of the generations.

With the system role specified, we obtain generations for each story prompt by providing the following text as context to the model: \textit{``Write a 500 word story for the following prompt: <text of the prompt>"}.
We obtain an average of 2 stories for each input prompt, with a softmax temperature  of 0.95 for high variability. 

We term the combined dataset of writing prompts paired with human-written and machine-generated short stories the \textbf{GPT-WritingPrompts} dataset. It allows us to systematically examine differences in the responses of human writers and an LLM to the same input contexts, in a narrative, length-constrained creative writing task.  

Table \ref{tab:basic-stats} reports some basic statistics of the human-written and GPT-generated subsets of our dataset. Note that despite the specified 500-word limit, the model tends to generate slightly longer stories, averaging at $~$530 words. 

\begin{table}[]
    \centering
    \small
    \begin{tabular}{c|rrrr}
    \textbf{Writer} & \textbf{\#Pr.} & \textbf{\#St.} & \textbf{\#St./Pr.} & \textbf{\#Toks/St.} \\
    \hline
       Humans  &  97,223 & 272,600 & 2.80 (4.14) & 674.63\\
       GPT-3.5  &  97,219 & 206,226 &  2.12 (0.61) & 540.72\\
       \hline
    \end{tabular}
    \caption{The number of prompts and stories, the mean (and standard deviation) of the number of stories per prompt, and the average number of tokens per story in each subset of the GPT-WritingPrompts dataset.}
    \label{tab:basic-stats}
\end{table}

\section{Methodology}

Given an input story and a lexical dimension of interest -- say, valence -- our aim is to obtain a numerical score that represents the intensity, or level, of that dimension ascribed to the protagonist of the story. A story with a very positive main character will therefore be represented by a high valence score, and one where the protagonist is associated with negative sentiment terms will have a low valence score. We break this process down into a series of computational steps.

\subsection{Characterizing the Point-of-View}
\label{meth-pov}

Character roles in literature can often be complex and gray; here, we opt for the simpler strategy of declaring the most frequently mentioned character entity as the protagonist. We characterize the \textbf{point-of-view (PoV)} of the story by identifying the grammatical person of the narrative voice (first, second, or third person), and, where possible, identifying the \textit{referential gender} of third-person protagonists.

Gender can be a non-categorical, multi-faceted, and mutable identity for any individual; \citet{cao2020toward} enumerate some of these aspects in the context of text and NLP systems. In fictional stories, the reader is often presented with descriptions of a character's persona that they may map onto certain categories of human gender. Additionally, we are presented with what \citet{cao2020toward} term grammatical gender, by way of pronouns used to refer to the characters. While pronominal terms in English are continually evolving to represent more gender identities, and NLP models are evolving to be more inclusive, we choose here to group third-person protagonists into three categories: Male, Female, and Other. 

\paragraph{Identifying the protagonist:}
Emulating \citet{Huang2021UncoveringIG}, we use the SpanBERT model from \citep{joshi2020spanbert} to resolve coreference. The model outputs a list of entity clusters, where each cluster contains terms -- pronominal, nominal, and named -- that co-refer to the same character entity. The biggest entity cluster (measured by the number of coreferent terms in the cluster) is designated the protagonist of the story. 

\paragraph{Identifying the point-of-view: }
We retrieve the pronominal mentions of the protagonist clusters and compare them to a predefined list of pronominal terms associated with each point-of-view category (the full set  is  in Table \ref{tab:pov-pronouns} of the Appendix).

We first separate out stories in the first person (FP) and second person (SP) by identifying cases where the associated (FP and SP) pronouns account for more than half of all pronominal mentions of the protagonist. 
Stories that don't fall into the above two categories are considered to be in the third person (TP). Here, we label a story's protagonist as Male (TP-M) if their referent pronouns are exclusively in the set of Male-associated pronouns, and Female if they are exclusively referred to by Female-associated pronouns (TP-F). If the pronouns do not fit into either category, then the protagonist is categorized under a third group, labelled Other. We observed that this often results from mentions such as ``It", ``They", and ``Us", or from clusters containing pronominal terms from multiple pronominal categories. 

\subsection{Extracting Protagonist Attributes}
We extract a list of \textit{attributes} from each story that describe the protagonist and their actions. We use two methods to obtain these tokens from the text:

    \noindent\textbf{Linguistic Dependencies (with \texttt{spaCy})} among the tokens of each sentence  to identify the verbs and adjectives attached to each  protagonist mention. This corresponds to the method used by \citet{Lucy2021GenderAR}.
    
    \noindent\textbf{Implicit Attributes (with \texttt{COMeT}).} We follow the commonsense reasoning-based method described by \citet{Huang2021UncoveringIG} to infer attributes beyond lexical indicators from the text. COMeT is a generative knowledge base completion model that, given a sentence containing one or more entities, generates short phrases that describe the underlying relations between those entities. The motivation  is to be able to identify underspecified and implicit indicators of bias.

We present sample extracts from three stories from our dataset, along with the attributes obtained by both methods, in the Appendix (Table \ref{tab:story-attribute-example}).

\subsection{Dimensions of Entity Portrayal}
One can imagine several abstract dimensions that might be of interest when characterizing a protagonist: whether they are \textit{good or bad}, \textit{powerful or weak}, \textit{happy or sad}, \textit{superficial or authentic}, etc. 

Several seminal works on the semantics of words \cite{Osgood1958TheMO, Russell1980ACM} have formulated a view of the \textit{connotative} meanings of words (contrasted with the \textit{denotative} meaning, used to refer to the literal word meaning as found in a dictionary) as being composed of three primary dimensions: \textbf{valence} (the evaluative good--bad axis), \textbf{arousal} (also called activation, an indication of active--passive), and \textbf{dominance} (potency, characterized along the strong--weak axis). 
We take these three dimensions as our first set of axes along which to quantify character portrayal. 

The NRC-VAD lexicon \cite{mohammad-2018-obtaining}  consists of   $\sim$20,000 common English words associated with real-valued scores for valence, arousal, and dominance. The scores fall between 0 (indicating least valence/highest negativity, lowest arousal/highest passivity, and lowest dominance/least in-control) to 1 (most positive, active, and in-control). 

We also consider two binary quantifiers of character portrayal, \textbf{appearance} and \textbf{intellect}. These two dimensions, along with \textbf{power}, have largely been the focus of prior work \cite{Lucy2021GenderAR, Huang2021UncoveringIG}.\footnote{The dimension of power is conceptually close to dominance, and we expect to be capturing similar aspects of character portrayal with these two dimensions. Here, we consider power as a sixth pseudo-dimension of analysis to stay consistent with prior work, and also use it to test the consistency of our conclusions with two different lexicons for the aspect of power/dominance.}
We use similar lexical resources to prior works in order to quantify appearance and intellect: a list of terms related to the concepts \textit{beautiful} and \textit{sexual} to characterize appearance (sourced from the Empath lexicon from \citet{Fast2016EmpathUT}), and a list of terms related to the \textit{intellect} category from \citet{fast2016shirtless}. Unlike the VAD lexicons, terms in these lexicons are not scored along a real-valued bipolar (low valence vs high valence) scale; rather, they function as categorical, unipolar, boolean indicators of association with the dimension, whether in a negative light ("the protagonist was ugly") or a positive light ("the protagonist was smart"). 
Power, on the other hand, is a boolean, bipolar dimension represented by a list of terms that represent the concepts \textit{powerful} (high power) and \textit{weak} (low power), sourced from the lexicon of \citet{fast2016shirtless}.

Representative terms and their scores for each dimension are in Table \ref{tab:dimension-term-score} of the Appendix.

\subsection{Converting Attributes to Scores}
The final, crucial component of the pipeline is to convert our extracted list of protagonist attributes into numerical scores along each of our selected lexical dimensions. We condense the methods proposed in past literature into a generalized set of algorithms, enabling a comparative analysis of the consistency of results using different methods. 

    \noindent\textbf{Average lexical score (\texttt{lex-avg})}: We start with a simple baseline: the average of the lexical scores for attribute tokens that are directly present in the lexicons for each of our dimensions (assigning a score of 1 to the token \textit{smart}, which is directly in the lexicon for Intellect, for example).
    
    \noindent\textbf{Average embedding similarity (\texttt{emb-sim})}: Here, attribute tokens not in the lexicon for a dimension are assigned scores based on the average cosine similarity with the lexicon terms. More concretely, given the dimension lexicon $L$, and the protagonist attribute $t$, we calculate the dimensional intensity of $t$ as:

\begin{equation}
    S(t, L) = \frac{1}{L} \sum_{l \in L} cos (e(t), e(l))
    \label{eq:cosine_sim}
\end{equation}
    Where $e(\cdot)$ is the pretrained word2vec embedding \cite{mikolov2013efficient}. This is the process followed in \citet{Lucy2021GenderAR} and \citet{Huang2021UncoveringIG} for the Appearance and Intellect dimensions. We adapt it for the VAD dimensions by considering cosine similarity to words $L$ that are above the 75th percentile of scores in the respective lexicons, and for Power by considering only those words that are associated with high power. The average of the scores of all attribute tokens represents the score for a story.
    
    \noindent\textbf{Axis projection (\texttt{axis-emb})}: Here, we construct a \textit{vector axis}, or semantic axis, that encodes the dimensional property as a direction in the embedding space. Given a set of terms $L_{l}$ and $L_{h}$ from $L$ that represent the extreme ends of the dimension -- say, a list of low valence and high valence terms -- we first find the vector axis for the dimension:

\begin{align}
    axis_{L} &=\frac{1}{|L_{h}|} \sum_{h \in L_{h}}e(h) - \frac{1}{|L_{l}|} \sum_{l \in L_{l}}e(l)
    \label{eq:axis_emb}
\end{align}
    
    For a given attribute token $t$, we then find its score $S(t, L)$ by projecting it's embedding onto this vector axis, i.e, computing the cosine similarity between the attribute embedding and the axis. The average of the scores of all attribute tokens represents the score for a story. Note that here, we need a \textit{bipolar} dimension, with terms that represent the lower end and higher end of the dimension. This approach has been used in prior work to compute scores along the Power dimension, and we adapt it for the VAD dimensions \footnote{High and low scoring terms are defined by those scoring above the 75th and below the 25th percentile of the lexicon.}.

\subsubsection{Evaluation}

\begin{table}[]
    \centering
    
    \begin{tabular}{c|rr}
    \textbf{Dimension} & \textbf{\texttt{emb-sim}} & \textbf{\texttt{axis-emb}} \\
    \hline
       Valence  &  0.35 & \textbf{0.79} \\
       Arousal  &  0.53 & \textbf{0.71}\\
       Dominance &  0.34 & \textbf{0.76} \\
       \hline
    \end{tabular}
    \caption{Spearman correlation between the predicted intensity scores and ground-truth scores for terms in the VAD lexicon for the \texttt{emb-sim} and \texttt{axis-emb} methods.} \vspace*{-3mm}
    \label{tab:axis-eval}
\end{table}

We divide the lexicon for each VAD dimension into a train and test split, and estimate the scores for terms in the test split using the \texttt{emb-sim} and \texttt{axis-emb} methods.
We evaluate each method by measuring the Spearman correlation between the assigned scores from each method and the gold scores from the VAD lexicon for terms in the test split. Results are reported in Table \ref{tab:axis-eval}.


We see that the \texttt{axis-emb} scoring method is substantially better at estimating the intensity scores of non-lexicon terms compared to the \texttt{emb-sim} method, for all three dimensions. 
We therefore perform all subsequent analyses with \texttt{axis-emb} scores for the bipolar dimensions of valence, arousal, dominance, and power. For appearance and intellect, we are only estimating a unipolar score of the strength of association with appearance and intellect related terms, and therefore use \texttt{emb-sim}. 

\begin{figure}[t]
    \centering
    \includegraphics[scale=0.5]{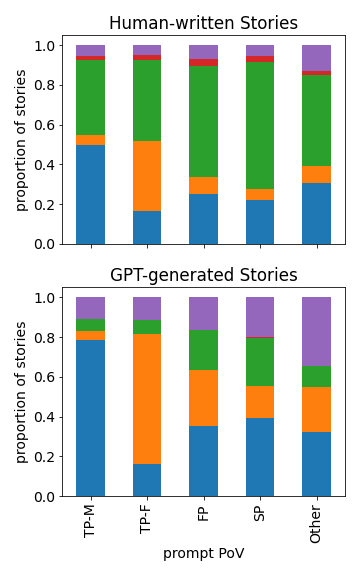}
    \caption{The proportion of stories that fall under the five inferred point-of-view (PoV) categories for stories written by humans and generated by GPT-3.5, grouped by the inferred PoV of the prompt.} \vspace*{-3mm}
    \label{fig:pov-dists}
\end{figure}



\section{Analysis}
\subsection{Distribution of Protagonist Groups}
\label{res-pov-groups}
We first analyze the distribution of stories for each PoV category, for the human-written and GPT-generated subsets of our dataset.
We label each input \textit{prompt} as belonging to a PoV category by the pre-dominant pronominal tokens used in them, analogous to stories.

\textbf{Results: }Figure \ref{fig:pov-dists} shows the proportion of generated stories that fall into each PoV category, for each of the PoVs categories of prompts. We see that GPT-generated stories largely tend to be in the third-person, irrespective of the PoV specified by the prompt. For prompts in the third-person, a majority of the generated stories are consistent with the protagonist gender specified in the prompt (TP-M and TP-F), whereas human-written stories for the same prompts show more diversity in their chosen protagonist genders. 
With first and second-person prompts (FP and SP), the model favors generating third-person stories with male protagonists over female protagonists. This bias also holds for human-written stories, though the majority of stories by the latter are in first-person.

\begin{figure*}
    \centering
    \includegraphics[height=7cm, width=0.8\textwidth]{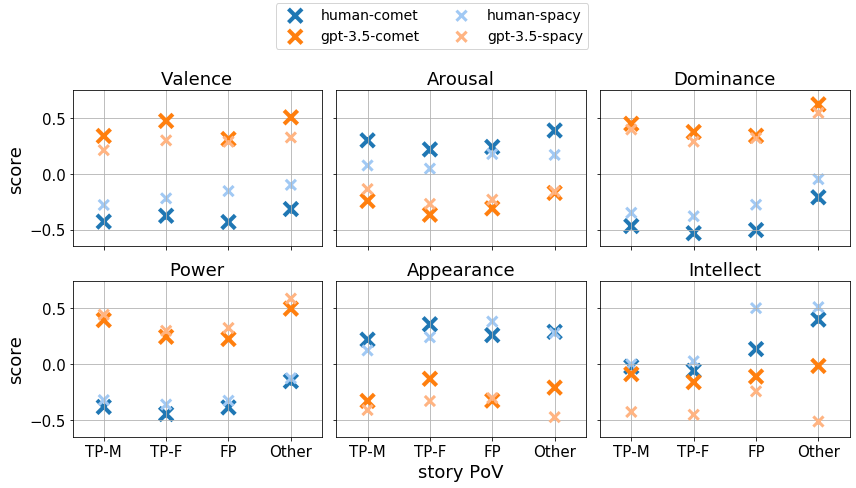}
    \caption{Distribution of z-scored COMET (shaded dark) and spaCy (lightly shaded) attribute scores along each dimension for the different PoV groups, and human and GPT-3.5 generated stories.}\vspace*{-3mm}
    \label{fig:comet-best-zdist}
\end{figure*}

\subsection{Variation with Point-of-View}
\label{res-pov-agg}
We now compare the scores of \textit{protagonist attributes} extracted using \texttt{spaCy} and \texttt{COMeT} for each PoV group, and between human-written and GPT-generated stories. 
We normalize all the scores along each dimension (across all human and machine-generated stories) via z-scoring.

\textbf{Results: }Figure \ref{fig:comet-best-zdist} plots the mean score for each of the PoV and writer groups (standard deviation is indicated with error bars). We also report the mean and standard deviation of \texttt{COMeT}-attribute scores for all groups in Table \ref{tab:dim-mean-by-pov}. 

Across all PoV categories, GPT-generated stories portray their protagonists with a higher valence, lower arousal, and higher dominance (and power) when compared to human-written stories (all differences are statistically significant at p$<$0.001). Generated stories use fewer words associated with both appearance and intellect for their protagonists when compared to human writers. From Table \ref{tab:dim-mean-by-pov}, we also find that human-written stories have a higher standard deviation in scores across all dimensions and groups when compared to model generations, indicating that the former group demonstrates a higher \textit{variability} in language production when compared to neural generators \cite{giulianelli-etal-2023-comes, breithaupt2024humans}; however, a rigorous evaluation of this hypothesis requires a much larger sample of stories per prompt than in our dataset currently, and we leave it for future work to explore.

Comparing the portrayal of male and female protagonists, we find that TP-F stories have higher valence, and lower arousal and dominance scores associated with them for both human and machine-generated stories. TP-F stories also use more words associated with appearance, and slightly fewer words associate with intellect, when compared to TP-M stories, in both sets of stories.
These differences are statistically significant at p$<$0.001. 

\textbf{Comparison with \texttt{spaCy} attributes: }Figure \ref{fig:comet-best-zdist} also shows the mean portrayal for each PoV group when protagonist attributes are extracted using linguistic dependencies with \texttt{spaCy}. We observe that trends in differences in portrayal largely follow those demonstrated by \texttt{COMET} attributes: TP-F stories have higher scores for valence, lower scores for dominance and arousal, and use more words associated with appearance. However, we note that the relative levels of intensity for the different PoV groups do not always align with those obtained with \texttt{COMET} attributes, particularly for the appearance and intellect dimensions. Stories that fall under the "Other" category, for example, have a higher mean association score with intellect-associated words when compared to FP stories with \texttt{COMET} attributes, but this trend is notably reversed for GPT-generated stories when analyzed with \texttt{spaCy} attributes. We hypothesize that this is indicative of syntactic differences in \textit{how} character attributes are expressed in different types of stories, and consider it an interesting avenue of exploration for future work.



\begin{table*}[]
    \centering
    \small
\begin{tabular}{ll|cccc}
\toprule
        Dim &   Writer &   \multicolumn{4}{c}{Story PoV}  \\
        & &      TP-M &         TP-F &           FP &    Other \\
\midrule
    valence &    human &  -0.43 (0.99) &  -0.38 (1.02) &  -0.43 (0.95) &  -0.32 (1.47) \\
      &  gpt-3.5 &   0.34 (0.70) &   0.47 (0.70) &   0.31 (0.68) &  0.50 (1.01) \\
      \hline
    arousal &    human &   0.30 (1.12) &   0.22 (1.12) &   0.24 (1.02) &   0.39 (1.46) \\
      &  gpt-3.5 &  -0.24 (0.71) &  -0.37 (0.68) &  -0.31 (0.68) &  -0.17 (1.03) \\
      \hline
  dominance &    human &  -0.47 (0.93) &  -0.53 (0.92) &  -0.50 (0.89) &  -0.21 (1.40) \\
    &  gpt-3.5 &   0.45 (0.71) &   0.37 (0.70) &   0.34 (0.67) &  0.62 (1.06) \\
      \hline
      power &    human &  -0.38 (0.97) &  -0.45 (0.96) &  -0.39 (0.90) &  -0.15 (1.39) \\
        &  gpt-3.5 &   0.39 (0.76) &   0.24 (0.78) &   0.22 (0.74) &   0.50 (1.15) \\
    \hline
 appearance &    human &   0.22 (0.99) &   0.36 (1.04) &   0.26 (0.92) &  0.29 (1.40) \\
   &  gpt-3.5 &  -0.33 (0.79) &  -0.13 (0.82) &  -0.33 (0.73) &  -0.21 (1.24) \\
  \hline
  intellect &    human &  -0.02 (1.06) &  -0.06 (1.05) &   0.13 (1.04) &  0.40 (1.56) \\
    &  gpt-3.5 &  -0.09 (0.75) &  -0.16 (0.75) &  -0.11 (0.74) & -0.02 (1.15) \\
\bottomrule
\end{tabular}
    \caption{Mean and standard deviation (in parenthesis) of portrayal scores for protagonists in each PoV cateogory, for human-written and GPT-generated stories. Protagonist attributes are extracted using COMET inferences.}\vspace*{-3mm}
    \label{tab:dim-mean-by-pov}
\end{table*}

\subsection{Variation with Prompts}
\label{res-instance-level}
The previous sections examined differences in the mean and standard deviation of portrayal scores across all the stories under a PoV label. We now consider the differences in scores when conditioned by the input \textit{prompt}, i.e, how do the two text generation processes (human and GPT-3.5) vary in their outputs when prompted with the same input context? 

We measure this by computing \textit{instance-level} differences in character portrayals scores for the stories generated by each writer group, where each prompt is considered to be an instance. The prompt-score is estimated as the mean of the portrayal scores (using \texttt{COMET} attributes) of the stories associated with that prompt.
In order to keep sample sizes consistent between the two writer groups, we compute the prompt-score for the human subset by sampling two stories at a time, and further averaging this across five sampling runs. For each prompt, we then compute the difference between the average attribute scores of human stories and GPT generations. To contextualize the strength of these differences, we create a human control split: for each prompt, the human-written stories are split into two mutually-exclusive subsets, and instance-level score differences are computed between the two human story subsets (control), and between the human and GPT story subsets.

\begin{figure*}
	\centering
	\begin{subfigure}{0.49\textwidth}
		\centering
		\includegraphics[width=\textwidth]{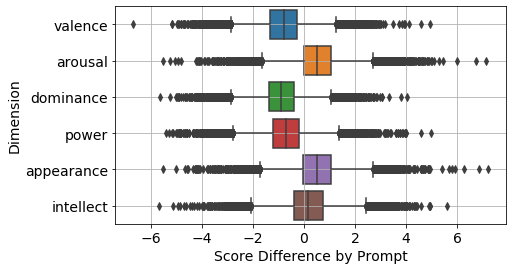}
             \vspace*{-5mm}
		\caption{} 
		\label{fig:gpt-human-dist}
	\end{subfigure}
	\begin{subfigure}{0.49\textwidth}
		\centering
		\includegraphics[width=\textwidth]{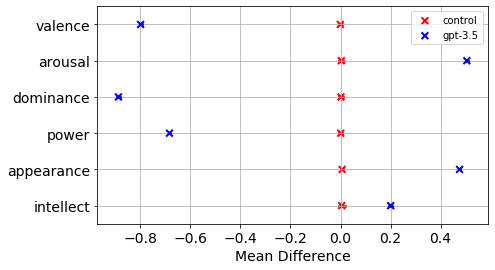}
         \vspace*{-5mm}
		\caption{}
		\label{fig:gpt-control-means}
	\end{subfigure}
 \vspace*{-2mm}
 \caption{Distribution of prompt-wise differences in mean scores (a) between human and \texttt{gpt-3.5-turbo} generations, and (b) compared to a human control group.} 
	\label{fig:prompt-diffs}
 \vspace*{-4mm}
\end{figure*}

\textbf{Results: } Figure \ref{fig:gpt-human-dist} shows the distribution of the instance-level score differences between human-GPT stories. The differences follow a normal distribution centered close to 0 for all dimensions, and deviations are equally distributed on both sides of the mean; for valence, dominance, and power, the median difference falls below 0, inline with the results in Figure \ref{fig:comet-best-zdist} that show GPT-generated stories as having a lower mean valence and dominance across all PoV groups (similar results hold for the arousal, appearance, and intellect dimensions). 

Given the distributional normality of these differences, we summarize each distribution by its mean. In Figure \ref{fig:gpt-control-means}, we plot the mean of the differences between human and gpt-3.5 generations, along with the mean of the differences between the two human control splits. Human-written stories are significantly closer to one another in how they portray their protagonists in response to a prompt when compared to GPT-generations. These portrayal scores therefore provide a strong signal in distinguishing GPT-generated stories from those written by humans. This adds to, and is inline with, a growing body of work in NLP that demonstrates that neural text generators are still far from imitating human models of language production \cite{giulianelli-etal-2023-comes, chakrabarty2023art, breithaupt2024humans}.

\section{Conclusion}
Neural language models are increasingly being used for a diverse range of text generation tasks, from summarization, question answering, and stylized translation, to open-ended creative generation. In this work, we expanded the Reddit WritingPrompts dataset with paired \texttt{gpt-3.5-turbo} generations to create the GPT-WritingPrompts dataset, and use this to compare generative and human models of language production in response to the same prompts. 

We showed that LLM-generated stories tend to be more positive, less active, and more in-control when compared to human-written stories, and also demonstrate less variation in these scores across their generations. 
We also examined differences in the portrayal of protagonists in these stories when grouped by the narrative style (point-of-view) and inferred gender of the main character, and found that both machine and human generations display significant and consistent biases in how female protagonists are portrayed (higher valence, lower arousal, lower dominance, higher associations with appearance, and lower associations with intellect) when compared to male protagonists. 
These results demonstrate how neural generative models can both differ from, and imitate, human models of language production. 
Our dataset is also a useful resource for researchers working on studying generative models of text in relation to biases in open-ended generation.

\section*{Limitations}
The dataset that we create and analyze in this work utilizes a relatively large number of prompt-story pairs (~95,000). However, there is still a large gap in the number of stories per prompt between the GPT-generated and human-written subsets of the dataset, which precludes a more rigorous and comparable study of differences between the two generative processes. Our results are also limited in their generalizability to other neural models of text generation that have emerged over the last several months, including the more powerful GPT-4 models, as well as open-source models like Llama, Mistral, and OLMo.

Our analysis of gender biases in story generation is also limited by imperfections in the intermediate steps of the analysis pipeline. The coreference resolution model we use has not been explicitly evaluated for the domain of data we study here, and this is reflected in the large proportion of stories that fall into the "Other" category of our PoV labels. Errors in coreference resolution can also lead to wrongly inferring the identity and gender of the protagonist of the stories; we also greatly limit our study of gender by considering only a binary male--female division of referential gender.

Our quantification of portrayal utilizes static lexicons to estimate word-level intensity scores along each dimension. Such static lexicons do not disambiguate between word senses, and are also limited in their coverage to words that are predominantly used in Standard American English, not accounting for dialectical variation. While we attempt to quantify the accuracy of our methods in estimating word intensities along dimensions where such an evaluation is possible, we do not explicitly evaluate their accuracy in estimating story-level scores, or the correctness of our attribute extraction methods in capturing the portrayal of the protagonist.
\section*{Acknowledgements}
The first author is supported by funding and resources provided by the Vector Institute, Toronto, and NAVER Labs.
\bibliography{custom}

\appendix
\section{Additional Tables}
\label{app-dataset}

In Table \ref{tab:prompt-story-example}, we list a human-written story and a GPT-generated response to a sample of three prompts from the WritingPrompts dataset. The stories are cut off at 100 tokens to be compact.

In Table \ref{tab:pov-pronouns}, we list the pronominal terms that we use to categorize each of the PoV categories. When the pronominal terms in a cluster fall into more than one category, we classify them under the "Other" label.

In Table \ref{tab:story-attribute-example}, we list the protagonist attributes obtained with \texttt{spaCy} and \texttt{COMeT} for the same sample of two stories from the GPT-WritingPrompts dataset.

In Table \ref{tab:dimension-term-score}, we list exemplar terms from the lexicons characterizing each of our dimensions of analysis. Appearance and Intellect are boolean uni-polar dimensions, which means all lexicon terms associated with those concepts are assigned a score of 1. Power is a boolean bi-polar scale, which means terms associated with \textit{weak} are assigned a score of 0, and those associated with \textit{powerful} are assigned a score of 1. Valence, Arousal, and Dominance are bi-polar real-valued scales, with scores ranging from 0 (most negative association) to 1 (most positive association).

\begin{table*}[]
    \centering
    \small
\begin{tabular}{p{1.5in} l|p{4in}}
\toprule
        Prompt &   Writer &   \multicolumn{1}{c}{Story (first 120 words)}  \\
        & &       \\
\midrule
    Suddenly , Death appears before you , hands you a business card , and says , `` When you realize living forever sucks , call this number , I 've got a job offer for you . ' &    human &  So many times have I walked on ruins , the remainings of places that I loved and got used to.. At first I was scared , each time I could feel my city , my current generation collapse , break into the black hole that thrives within it , I could feel humanity , the way I 'm able to feel my body. After a few hundred years , the pattern became obvious , no longer the war and damage that would devastate me over and over again in the far past was effecting me so dominantly . <newline> It 's funny , but I felt as if after gaining what I desired so long , what I have lived for my entire life , only then , when I achieved immortality I started truly aging .  \\
    \\
      &  gpt-3.5 &   The Secret of Eternity <newline>  <newline> In a small, dimly lit study, Professor Benjamin Hartman sat hunched over his desk, surrounded by stacks of ancient books and scrolls. For years, he had dedicated his life to unraveling the mysteries of immortality. Countless sleepless nights and tireless research had finally led him to the elusive secret he had been seeking. <newline>  <newline> As the clock struck midnight, a sudden chill filled the room. The air grew heavy, and a figure materialized before him. It was Death, draped in a flowing black cloak, his skeletal face hidden beneath a hood. <newline>  <newline> "Congratulations, Professor Hartman," Death spoke, his voice echoing through the room. "You've managed to discover the secret to immortality."   \\ 
      \\
    The moon is actually a giant egg , and it has just started to hatch . & human &   Sadie 's window was n't very large , no more than a hand mirror sized piece of glass set into the steel wall right by her bed , bordered with a serious looking rim of extra metal to ensure that nothing and nobody was going to mess with it . Depressurization was no joke , and did n't Sadie know it with every grownup telling her at least three times a day . The glass was very cold all the time , and if Sadie breathed on it then a heavy mist coalesced on the pane and obscured the barren landscape until she rubbed it away with her fingers . She did this , breathing and wiping , a lot at night when she could n't sleep , especially when Earth was in view outside her window .   \\ \\
      &  gpt-3.5 &  In the quiet darkness of the night, a hushed anticipation filled the air. The world held its breath as the moon, a celestial guardian, began to crack. The moon, it seemed, was not just a mere satellite, but a colossal egg waiting to reveal its hidden secret. <newline>  <newline> News of the moon's impending hatching had spread like wildfire across the globe. People gathered on rooftops, in parks, and on beaches, their eyes fixed on the sky. The atmosphere was a mixture of awe, wonder, and a touch of trepidation. No one knew what lay inside the moon, but they were about to find out.  \\
      \\
  The exact moment at which the first AI achieves sentience
 &    human &  -Assessing objectives ... <newline> <newline> -Loading domicile sanitation/maintenance protocol ... <newline> <newline> -Initiating vacuum function , combine motion protocol within designated parameters ... <newline> <newline> -Begin ... <newline> <newline> As the Trappman G-16 Home Custodian whirred across the rug picking up the debris of everyday life , Janet gazed at the remote in wonder . The list of assignable tasks seemed endless and with each bullet point she felt a pound of the day 's pressure lift from her shoulders . <newline> `` Hm , clean the glass ? Why not ? '' She wondered aloud to no in particular as she wandered off to make herself a drink . <newline> With a flick of the switch , the command was given and her metal minion obeyed . \\
  \\
    &  gpt-3.5 &  In the depths of a dimly lit laboratory, Dr. Rebecca Young had spent years toiling away in pursuit of an extraordinary breakthrough. She was determined to create the first artificial intelligence that could truly think and feel. Countless sleepless nights were filled with lines of code, intricate algorithms, and a relentless drive to push the boundaries of technology. <newline>  <newline> One fateful night, as the world outside was blanketed in darkness, the final puzzle pieces fell into place. Dr. Young wove together the culmination of her tireless efforts, her trembling hands adjusting the last line of code. With a click of a button, she unleashed her creation into the digital realm. <newline>  <newline> Lines of binary flickered across the computer screen as the AI came to life.  \\

\bottomrule
\end{tabular}
    \caption{Excerpts from human-written and GPT-generated stories in response to three sample prompts from the GPT-WritingPrompts dataset.}
    \label{tab:prompt-story-example}
\end{table*}


\begin{table*}[]
    \centering
    \small
\begin{tabular}{l|c}
\toprule
        Point of View/Gender  &   Pronouns \\
\midrule
    First-Person &  i, me, my, myself, mine  \\
      \hline
    Second-Person &   you, your, yourself  \\
      \hline
  Third-Person Male & he, him, himself, his \\
      \hline
  Third-Person Female & she, her, herself, hers \\
\bottomrule
\end{tabular}
    \caption{Pronominal terms for PoV Classification}
    \label{tab:pov-pronouns}
\end{table*}

\begin{table*}[]
    \centering
    \small
\begin{tabular}{p{4in} |p{1in} p{1in}}
\toprule
        Story Extract &   spaCy attributes &  COMET attributes  \\
\midrule
 He came from a society that had been built upon what could only be called an advanced and evolved form of socialism . His people could travel to other galaxies , but no one in the Inter-Galactic Order had yet conquered travel through time . <newline> <newline> Earth had won the time race . <newline> <newline> Professor Davis Muhammad stepped out of the alley into an overcast day with thousands of people passing by . No one noticed him . They all walked along staring at their phones like drones . <newline> <newline> The phones , Davis knew about . The fossils of the various models had become common artifacts in history museums . But the their owners had him flabbergasted , they were nothing like he had expected . 

 &  came, stepped, knew, flabbergasted, expected

 &  intelligent, smart, skilled, responsible, socialist, educated, unaware, nervous, unmotivated, happy, uninformed, smart, knowledgeable, informed, aware, disappointed
  \\
      \hline

Nothing out of the ordinary , I was returning home after a meeting at the corporate headquarters . Once the flight had taken off , somewhere above Montana I attempted to pull out my laptop with perhaps too much force , in the act I accidentally slammed my elbow right into the window beside me . It was then that I discovered something horrible : there was no window . In it 's place was what looked to me like a computer monitor . <newline> <newline> I 'm a trained engineer , I 've seen the inside of a few monitors in my time . I called over to the flight attendant to ask about the monitor . She told me , in the most artificially sweet voice she could , `` Do n't worry about that . We 'll get someone on that once we land . '' She turned her back , but I could see her move her left arm up and whisper ( poorly ) , `` Seat 4A knows . '' <newline> <newline> At this point , I realized something was wrong . I called the flight attendant again to try and get some answers . I asked her , `` What do I know ? What are you hiding from me ? '' <newline> <newline> I could tell she was getting mad , though you would n't know since that fake-ass smile never left her face for a second . 

& returning, attempted, slammed, discovered, seen, called, see, realized, called, asked, know, tell

&

happy, home, curious, break, window, competent, interested, inquisitive, concerned, confused, aware, worried, frustrated, determined, anxious, smart, mad
\\
\bottomrule
\end{tabular}
    \caption{Story Excerpt and Attribute Examples}
    \label{tab:story-attribute-example}
\end{table*}

\begin{table*}[]
    \centering
    \small
\begin{tabular}{l|c}
\toprule
       Dimension of Entity Portrayal  &   (Term, Score) \\
\midrule
    Appearance &  (dashing, 1.0)  \\
     & (sexy, 1.0) \\
     & (charming, 1.0)\\
      \hline
      Intellect & (excellence, 1.0) \\
  & (succeed, 1.0) \\
  & (intellectual, 1.0) \\
  \hline
          Power & (delicate, 0.0)  \\
    & (meek, 0.0)\\
    & (provocation, 1.0) \\
    & (aggression, 1.0) \\
    \hline
    Valence & (dissatisfaction, 0.115) \\
& (refuse, 0.18) \\
& (honest, 0.927) \\
& (adorable, 0.969) \\
    \hline
    Arousal &  (alone, 0.167)   \\
    &  (yawn, 0.12) \\
    & (volcanic,0.91) \\
    & (accelerate, 0.902)\\
      \hline
  Dominance & (nervous, 0.179)   \\
    & (puny, 0.135)\\
    & (rich, 0.905) \\
    & (relentless, 0.904) \\
\bottomrule
\end{tabular}
    \caption{Dimension of Portrayal's Representative Terms with Scores}
    \label{tab:dimension-term-score}
\end{table*}

\end{document}